\documentclass{article}

\usepackage{spconf}
\usepackage{latexsym}
\usepackage{amsmath}
\usepackage{amssymb}
\usepackage{graphicx}
\usepackage{booktabs}
\usepackage{multirow}
\usepackage{diagbox}
\usepackage{enumitem}
\usepackage{hhline}
\usepackage[numbers, sort&compress]{natbib}

\newcommand{\Tlab}[1]{\label{table:#1}}
\newcommand{\Tref}[1]{Table \ref{table:#1}}
\newcommand{\flab}[1]{\label{fig:#1}}
\newcommand{\fref}[1]{Fig. \ref{fig:#1}}

\begin{document}

\title{Explainable analysis of Deep Learning Methods \\ for SAR Image Classification}

\name{Shenghan Su\textsuperscript{1},
    Ziteng Cui\textsuperscript{1},
    Weiwei Guo\textsuperscript{2}*,
    Zenghui Zhang\textsuperscript{1},
    Wenxian Yu\textsuperscript{1}}

\address{
    \textsuperscript{1}Shanghai Key Laboratory of Intelligent Sensing and Recognition, Shanghai Jiao Tong University,\\ Shanghai, 200240, China\\
    \textsuperscript{2}Center of Digital Innovation, Tongji University, Shanghai, 200092, China
}

\maketitle

\begin{abstract}
   Deep learning methods exhibit outstanding performance in synthetic aperture radar (SAR) image interpretation tasks. However, these are black box models that limit the comprehension of their predictions. Therefore, to meet this challenge, we have utilized explainable artificial intelligence (XAI) methods for the SAR image classification task. Specifically, we trained state-of-the-art convolutional neural networks for each polarization format on OpenSARUrban dataset and then investigate eight explanation methods to analyze the predictions of the CNN classifiers of SAR images. These XAI methods are also evaluated qualitatively and quantitatively which shows that \textit{Occlusion} achieves the most reliable interpretation performance in terms of Max-Sensitivity but with a low-resolution explanation heatmap. The explanation results provide some insights into the internal mechanism of black-box decisions for SAR image classification.

\end{abstract}

\begin{keywords}
   SAR Image Classification; Explainable Artificial Intelligence; Deep Learning; Sentinel-1; OpenSARUrban
\end{keywords}

\section{INTRODUCTION}

Deep learning methods are playing an increasingly remarkable and irreplaceable role in practical applications of various fields~\cite{DeepLearningMethodsandApplications},  including the field of synthetic aperture radar (SAR)~\cite{PolarimetricSARImageClassificationUsingDeepConvolutionalNeuralNetworks}. In order to fit complex data distribution and obtain better performance, they generally have an  intricate network structure of opacity and extreme non-linearity. However, from a more rigorous perspective, the application of deep learning models in many fields is still limited, for lacking more reliable interpretability and explainability~\cite{InterpretabilityOfDeepLearningASurvey}. 

To circumvent these limitations, various explainable artificial intelligence (XAI) methods have been proposed, which provide some detailed explanations for the  decision-making process of deep neural networks. The XAI methods could analyze and reveal the internal mechanisms to some extent, so as to give corresponding explanations and further evaluate the deep learning models~\cite{UnmaskingCleverHansPredictors}.

In the field of Remote Sensing (RS), Ioannis et al. recently employed the XAI methods to analyze the deep learning models for optical remote sensing image classification based on the BigEarthNet and SEN12MS datasets~\cite{EvaluatingXAI}. Nevertheless, there is little work of explainable analysis of the deep neural networks for SAR image classification, despite their success in it.

In this paper, we conducted an explainable analysis for deep convolutional neural networks (DCNNs) for SAR image classification. Our contributions are:

\begin{itemize}[topsep=0mm,itemsep=0mm,parsep=0mm,leftmargin=5mm]

    \item We train a Residual Neural Network~\cite{He_2016_CVPR} ResNet101 on  the OpenSARUrban dataset~\cite{opensarurban}, achieving state-of-the-art performance.

    \item We employ eight XAI methods to explain the predictions of the ResNet101 network. We also evaluate these XAI methods qualitatively and quantitatively.

    \item We conduct an in-depth analysis of the experimental results and draw some  inspiration from explainable SAR image classification.
    
\end{itemize}

\section{methodology}
\begin{table*}
    \centering
    \caption{Results of quantitative metrics.}
    \renewcommand\arraystretch{1.1}
    \setlength{\tabcolsep}{3mm}{
        \begin{tabular}{|c|cc|cc|}
\hline 

\multirow{2}{*}{\diagbox{XAI Method}{Quantitative Metrics}} & \multicolumn{2}{c|}{Max-Sensitivity} & \multicolumn{2}{c|}{XAI entropy (KB)} \\ \cline{2-5} 
                            & \multicolumn{1}{c|}{VH}          & VV         & \multicolumn{1}{c|}{VH}         & VV         \\ \hline
Integrated Gradients        & \multicolumn{1}{c|}{0.0742}      & 0.0952     & \multicolumn{1}{c|}{28.663}     & 30.841     \\ \hline
Input × Gradient            & \multicolumn{1}{c|}{0.1362}      & 0.1527     & \multicolumn{1}{c|}{29.123}     & 29.812     \\ \hline
Guided Backpropagation      & \multicolumn{1}{c|}{0.0412}      & 0.0392     & \multicolumn{1}{c|}{24.34}      & 24.099     \\ \hline
Deconvolution               & \multicolumn{1}{c|}{0.0766}      & 0.0754     & \multicolumn{1}{c|}{29.638}     & 30.058     \\ \hline
Saliency                    & \multicolumn{1}{c|}{0.1376}      & 0.1534     & \multicolumn{1}{c|}{52.624}     & 52.14      \\ \hline
Occlusion                   & \multicolumn{1}{c|}{0.0178}      & 0.0195     & \multicolumn{1}{c|}{4.305}      & 4.191      \\ \hline
Guided Grad-CAM             & \multicolumn{1}{c|}{0.1819}      & 0.1824     & \multicolumn{1}{c|}{11.682}     & 11.451     \\ \hline
Grad-CAM                    & \multicolumn{1}{c|}{0.4528}      & 0.4942     & \multicolumn{1}{c|}{15.107}     & 14.916     \\ \hline
\end{tabular}
    }
    \Tlab{XAI_result}
\end{table*}
\subsection{XAI Methods}
We firstly train a ResNet101\cite{He_2016_CVPR} on the OpenSARUrban dataset\cite{opensarurban} which achieves state-of-the-art performance,  and then apply eight XAI methods to understand the internal behavior of the SAR image classifier, including \textit{Saliency}~\cite{Saliency}, \textit{Input × Gradient}~\cite{InputXGrad}, \textit{Integrated Gradients}~\cite{IntGrad}, \textit{Guided Backpropagation}~\cite{GuidedBackprop}, \textit{Grad-CAM}~\cite{GradCAM}, \textit{Guided Grad-CAM}~\cite{GradCAM} and \textit{Deconvolution}~\cite{Deconvolution} .

The \textit{Input × Gradient (InputXGrad)} method multiplies the input with the gradient by Equation~\ref{InputXGrad_eq}. Compared with using the gradient alone, it shows more reliability~\cite{EvaluatingXAI}.
\begin{equation}\label{InputXGrad_eq}
\Phi_{\text {InputXGrad}}\left(f_{c}, x\right)  =  x \odot \nabla f_{c}(x)
\end{equation}

The \textit{Guided Backpropagation} method computes the gradient of the target output with respect to the input, but only non-negative gradients are backpropagated in that the gradients of ReLU functions are overridden.

The \textit{Integrated Gradients (IntGrad)} method represents the integral of the gradients along the straightline path from a base input $x^{\prime}$ to the input $x$\cite{IntGrad}. The base input can be a root point of the desired function, which, for instance, can be the black image for the networks.
\begin{align}\label{IntGrad_eq}
\footnotesize
\Phi_{\text{IG}}\left(f_{c}, x\right)=\left(x-x^{\prime}\right) \times &\left.\int_{0}^{1} \frac{\partial f_{c}(\tilde{x})}{\partial \widetilde{x}}\right|_{\tilde{x}=x^{\prime}+a\left(x-x^{\prime}\right)} d a 
\quad 
\end{align}

The \textit{Occlusion} is a perturbation based approach that replaces each rectangular region with a given baseline / reference, and compares their output differences.
 
\begin{align}\label{Occlusion_eq}
\Phi_{\text {Occlusion }}^{p_{i}}\left(f_{c}, x\right)=f_{c}(x)-f_{c}\left(x_{p_{i}}^{\prime}\right)\end{align}

\subsection{Evaluating Metrics}
In order to evaluate the performance of the XAI methods, quantitative evaluation and qualitative evaluation were adopted by the metrics \textit{Max-Sensitivity}~\cite{MaxSensitivity}  and \textit{XAI entropy}, namely,\textit{File Size}\cite{FileSize}.

The \textit{Max-Sensitivity}~\cite{MaxSensitivity} metric measures the stability on the basis of the maximum explanation change with a subtle input disturbance, as in Equation \ref{MaxSensitivity_eq}. The smaller \textit{Max-Sensitivity} score indicates a  more reliable explanation.
\begin{align}\label{MaxSensitivity_eq}
\operatorname{SENS}_{MAX}\left(\Phi, f_{c}, x, r\right) & = \max _{\left\|x^{\prime}-x\right\|  \leq r}\left\|\Phi\left(f_{c}, x^{\prime}\right)-\Phi\left(f_{c}, x\right)\right\|
\end{align}
The \textit{XAI entropy} metric is the auxiliary evaluation index that measures the amount of information a XAI visualization image contains. 

 \section{EXPERIMENTS}
 
\begin{figure*}
    \centering
    \includegraphics[width=165mm]{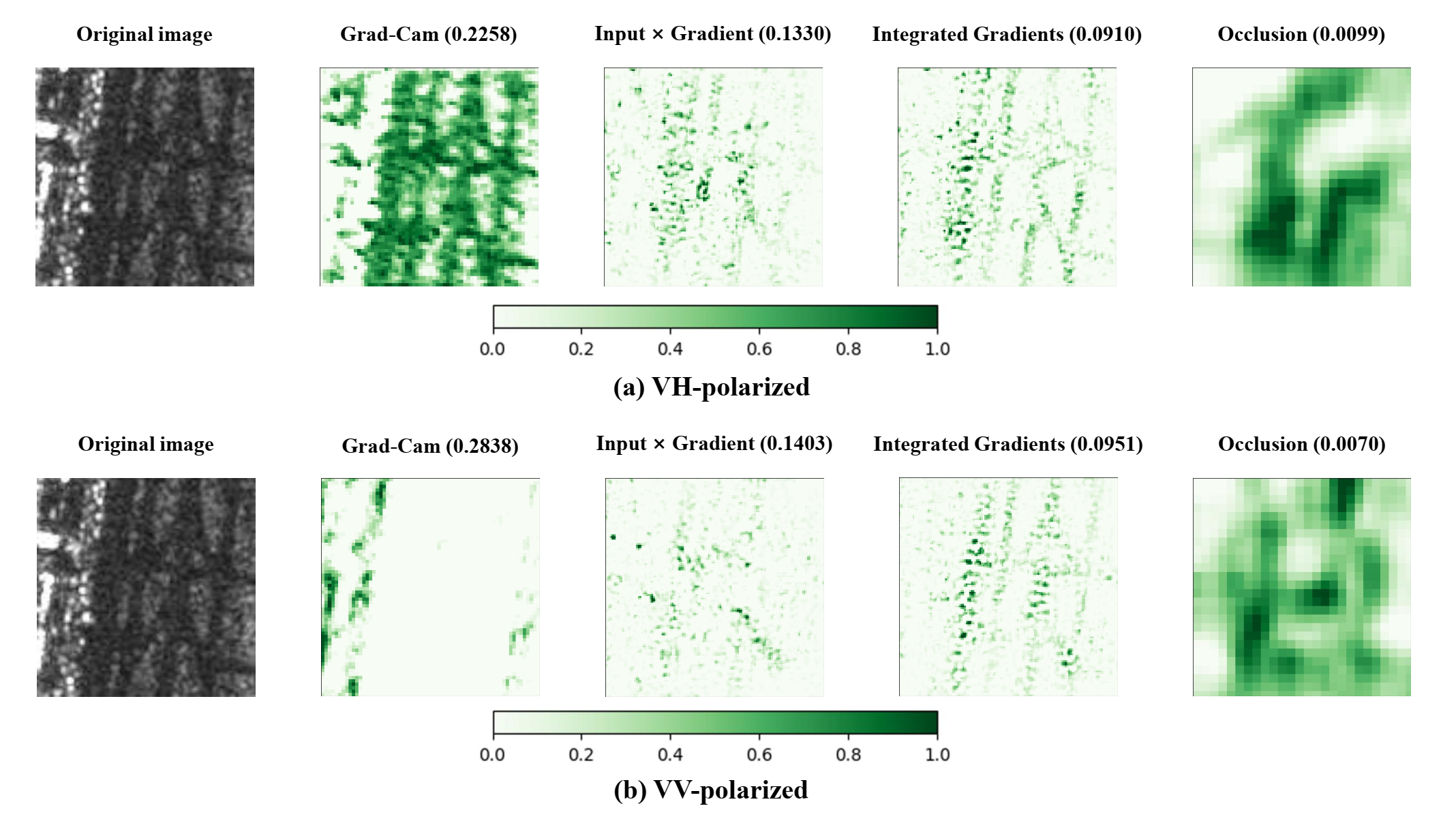}
    \caption{Qualitatively evaluating the predictions of ResNet101 for the airport class in the OpenSARUrban dataset. Max-Sensitivity score in parenthesis after method’s name.}
    \flab{Qualitative_Evaluation}
\end{figure*}

\subsection{Experimental Setup}

\subsubsection{Dataset}

Experiments were conducted on the OpenSARUrban dataset\cite{opensarurban}. It is built on the Sentinel-1 imagery and consists of 10 different urban categories, covering 21 major cities in China. Each image patch includes 2 kinds of polarization, VV and VH polarization.

\subsubsection{Implementation Details}

Due to the different polarization modes, we divide the OpenSARUrban dataset into two parts: one is the VH-polarized part and the other is the VV-polarized part. We conduct the following experiments on these two parts separately.

We firstly adopt the ResNet101 mode\cite{He_2016_CVPR} that achieves 83.44 percent classification accuracy in VH-polarized part and 84.44 in VV-polarized part of OpenSARUrban, by the data augmentation and integrating learning rate scheduler strategies. In the following, the XAI methods are utilized to comprehend the prediction of the trained models: \textit{Saliency}, \textit{Input × Gradient}, \textit{Integrated Gradients}, \textit{Guided Backpropagation}, \textit{Grad-CAM}, \textit{Guided Grad-CAM}, \textit{Occlusion}, \textit{Deconvolution}. We apply attribution analysis with visual examination and visualized the attributions by generating the heat maps for each image for every XAI method. At the same time, \textit{Max-Sensitivity} and \textit{XAI entropy} are applied to quantitatively evaluate the performance of the above XAI methods. The XAI experiments are implemented using the Captum library ~\cite{kokhlikyan2020captum}. 

\subsection{Results and Discussion}

\subsubsection{Quantitative Evaluation}

\Tref{XAI_result} exhibits the experimental results of different XAI methods on the urban scenes classification task in SAR images.

In terms of the \textit{Max-Sensitivity} metric, it is worth noting that lower scores indicate higher performance. Specifically, \textit{Occlusion}, \textit{Guided Backpropagation} and \textit{Integrated Gradients} are among the top three lowest scores in the VH-polarized part while \textit{Occlusion}, \textit{Guided Backpropagation} and \textit{Deconvolution} are among the three lowest scores in the VV-polarized part. On the other hand, \textit{Grad-CAM} attains the highest score in both polarized data. For all cases, \textit{Occlusion} is the most prominent XAI method with a \textit{Max-Sensitivity} score of less than 0.05, while \textit{Grad-CAM} performs the worst under \textit{Max-Sensitivity} metric in interpreting the SAR image classifier's prediction result.

In terms of the \textit{XAI entropy} metric, the ranking results are roughly the same in both polarized datasets. Specifically, \textit{Occlusion}, \textit{Guided Grad-CAM} and \textit{Grad-CAM} obtain the top three lowest scores in both polarized data. In addition, \textit{Occlusion} ( 15 × 15 sliding window with 5 × 5 strides) provides us with more low-resolution information and rough explanations in interpreting the SAR image classifier's prediction result, which possibly makes it easier for humans to understand this model.

\subsubsection{Qualitative Evaluation}

A representative case that we visualize the attributions of the XAI methods is presented in \fref{Qualitative_Evaluation}. Overall, after careful observation, all of the XAI methods manage to analyze and reveal the internal mechanism in both VH- and VV-polarized dataset. 

Concerning visual examination, \textit{Occlusion}, \textit{Integrated Gradients} and \textit{Input × Gradient} generally successfully mark the airstrip as an important area that corresponds to the airport class in both VH- and VV-polarized dataset, which is similar to our human cognition. However, it is worth mentioning that \textit{Grad-CAM} exhibits too much explanation of many details, resulting in poorer anti-interference and higher \textit{Max-Sensitivity} scores in the VH-polarized dataset. It seemed that less attention was focused on the airstrip in the VV-polarized dataset, leading to insufficient interpretability. 

\subsubsection{Discussion}
Regarding the performance of XAI methods, on the one hand,  \textit{Occlusion} and \textit{Integrated Gradients} are the most interpretable XAI methods in both quantitative evaluation and qualitative evaluation. They successfully explained the SAR image classification task by targeting the most relevant area of each image. Moreover, they achieved the lowest \textit{Max-Sensitivity} scores, which indicates that when the input slightly changes, the interpretation result has no obvious difference, resulting in the highest reliability. 

On the other hand, for quantitative evaluation, \textit{Grad-CAM} performed worst, which exhibited too much explanation of many details and ignored the main relevant regions of each image, resulting in poorer anti-interference and higher \textit{Max-Sensitivity} scores. Additionally, for qualitative evaluation, \textit{Deconvolution} was the least interpretable XAI method, although it achieved relatively low \textit{Max-Sensitivity} scores. The reason is 
that the internal mechanism of \textit{Deconvolution} causes gradients of ReLU functions to be overridden.

\section{CONCLUSIONS}
In this paper, we utilized eight XAI methods to explain the ResNet101's prediction of SAR image classification on OpenSARUrban and evaluate them qualitatively and quantitatively. We observe  that each method has its own merits. \textit{Occlusion} achieves the most reliable interpretation performance with the lowest \textit{Max-Sensitivity} score. However, it failed to provide high-resolution information and detailed explanations. The XAI methods can reveal the internal mechanism of deep learning methods and provide some insights into SAR image classification tasks to some extent. 

\setlength{\bibsep}{0.6ex}
\bibliographystyle{IEEEtran}
\bibliography{egbib}

\end{document}